\documentclass{article}

\usepackage{spconf,amsmath,graphicx}

\usepackage{times}
\usepackage{epsfig}
\usepackage{graphicx}
\usepackage{amsmath}
\usepackage{amssymb}
\usepackage{verbatim}
\usepackage{booktabs} 
\usepackage{adjustbox}
\usepackage{colortbl} 
\usepackage{xcolor} 
\usepackage{xfrac}
\usepackage[]{algorithm2e}
\usepackage{subcaption}
\usepackage{graphicx,multirow}
\usepackage{array}

\title{Generalized Bilinear Deep Convolutional Neural Networks for Multimodal Biometric Identification}
\name{Sobhan Soleymani, Amirsina Torfi, Jeremy Dawson, and Nasser M. Nasrabadi, {\it Fellow, IEEE}}
\address{West Virginia University}

\begin{document}
%
\maketitle
\begin{abstract}
\vspace{-1mm}
In this paper, we propose to employ a bank of modality-dedicated Convolutional Neural Networks (CNNs), fuse, train, and optimize them together for person classification tasks. A modality-dedicated CNN is used for each modality to extract modality-specific features. We demonstrate that, rather than spatial fusion at the convolutional layers, the fusion can be performed on the outputs of the fully-connected layers of the modality-specific CNNs without any loss of performance and with significant reduction in the number of parameters. We show that, using multiple CNNs with multimodal fusion at the feature-level, we significantly outperform systems that use unimodal representation. We study weighted feature, bilinear, and compact bilinear feature-level fusion algorithms for multimodal biometric person identification. Finally, We propose generalized compact bilinear fusion algorithm to deploy both the weighted feature fusion and compact bilinear schemes. We provide the results for the proposed algorithms on three challenging databases: CMU Multi-PIE, BioCop, and BIOMDATA. 
\end{abstract}
\vspace{-3mm}
\section{Introduction}
\vspace{-3mm}
The permanence and uniqueness of human physical characteristics such as face, iris, fingerprint, and voice is widely utilized in biometric systems deploying the corresponding feature representation of these characteristics~\cite{shekhar2014joint}. Multimodal biometric models have demonstrated more robustness to noisy data, non-universality and category-based variations~\cite{jaafar2013review,toli2014survey}. The multimodal networks can improve recognition task in cases where one or more of the biometric traits are distorted. A recognition algorithm using a multimodal architecture, requires selecting the discriminative and informative features from each modality as well as exploring the dependencies between different modalities. This architecture should also discard the single modality features that are not useful in joint recognition. 

However, employing a fusion algorithm is the most prominent  challenge in multimodal biometric systems~\cite{nagar2012multibiometric}. The fusion algorithm can be performed at signal, feature, score, rank or decision levels~\cite{connaughton2013fusion
} using different schemes such as feature concatenation~\cite{shi2016rule,
goswami2016group,Soleymani2018multi} and bilinear feature multiplication~\cite{lin2015bilinear,chowdhury2016one}. Although score-, rank- and decision-level fusion are studied in the literature extensively, since these levels are easier to access in the biometric systems, feature-level fusion results in a better discriminative classifier~\cite{
ross2005feature} due to the preservation of raw information~\cite{shekhar2014joint}. Feature level fusion integrates different features extracted from different modalities to a more abstract feature representation, which can further be used for classification, verification, or identification~\cite{
haghighat2016discriminant}.

To integrate the features from different modalities, several fusion methods have been considered~\cite{
shi2016rule}. The prevalent fusion method in the literature is feature concatenation, which is very inefficient exploiting the dependency between the modalities as the feature space dimensionality increases~\cite{nagar2012multibiometric,goswami2016group}. To overcome this shortcoming, bilinear multiplication of the individual modalities is proposed~\cite{lin2015bilinear,chowdhury2016one}. Using bilinear multiplication, the higher-level dependencies between the modalities are exploited and enforced through the backpropagation algorithm. The bilinear multiplication is effective since all of the elements of the single modalities interact through multiplication. The main issue in bilinear operation is the high dimensionality of its output regarding the cardinality of the inputs. Recently, to handle this shortcoming, compact bilinear pooling is proposed~\cite{gao2016compact,delbrouck2017multimodal,fukui2016multimodal}.

Convolutional neural networks are recently utilized for classification of multimodal biometric data. Although, CNNs are mainly used as classifiers, they are also efficient tools to extract and represent discriminative features from the raw data. Compared to hand-crafted features, employing CNN as domain feature extractors has demonstrated to be more promising when facing different biometric modalities such as face~\cite{lawrence1997face,kazemi2018attribute}, iris~\cite{gangwar2016deepirisnet} and fingerprint~\cite{nogueira2016fingerprint}.

In this paper, we make the following contributions: (i) instead of spatial fusion at the convolutional layers, modality-dedicated networks are designed to extract modality-specific features for the fusion; (ii) a fully data-driven architecture using fused CNNs and end-to-end joint optimization of the overall network, is proposed for joint domain-specific feature extraction and representation with the application of person classification; finally (iii) weighted feature fusion 
and generalized compact bilinear feature fusion are considered at the fully-connected level.  
\vspace{-3mm}
\section{Generalized compact bilinear fusion}\label{Bilibear}
\vspace{-3mm}
Consider a fusion operation $f: (X_1, X_2,..., X_n)\rightarrow Y$ that fuses $n$ modalities;
$X_i\in R^{H_i\times W_i \times D_i}$, $i=1,2,..,n$. The fusion operation results in $Y\in R^{\widehat{H}\times \widehat{W} \times \widehat{D}}$, where $W$, $H$ and $D$ correspond to width, height and depth of the feature maps. Fusion can be performed using the feature maps of the CNNs when the corresponding feature maps from different modalities are compatible. However, in multimodal biometric networks, the feature maps can vary in the spatial dimension due to the different spatial dimensionality of the inputs. To handle this issue, instead of utilizing convolutional layers feature maps for fusion, fully-connected layers are considered in our architecture for ultimate modality-dedicated feature representation. Therefore, in our proposed architecture, 
$H_i=W_i=\widehat{H}=\widehat{W}=1$, and there is no condition on $D$. We show that the fully-connected representation provides promising results in the case of recognition applications. 

In the proposed fusion algorithm, prior to the fusion, each modality is represented by the output of a fully-connected layer which we call the modality-dedicated embedding layer. In {\bf weighted feature fusion} algorithm, the fusion function concatenates the modality-dedicated embedding layers of the multiple modalities, in which $Y \in R^{1 \times \widehat{D}}$, where $\widehat{D}=\sum_i D_i$. In {\bf bilinear fusion} algorithm, $Y=X_1^T X_2$. If $H_i\neq 1$, the outer product is applied on two feature maps at the pixel level, followed by global average pooling over the spatial dimensions~\cite{lin2015bilinear,chowdhury2016one}. However, the bilinear fusion over fully-connected layers computes the outer product of the modality-dedicated embedding layers, where $Y \in R^{1 \times \widetilde{D}}$ and $\widehat{D}=\prod_i D_i$. The resulting feature-level representation $Y \in R^{1 \times \widehat{D}}$, projects all possible feature-level interactions between the $n$ modalities. In the case that the $n$ is larger than two, in each step the outer product is vectorized and then multiplied by the next modality.
\\
{\bf Generalized compact bilinear feature-level fusion algorithm}: Compact bilinear fusion projects the outer product of two vectors into a low-dimensional sub-space with very little loss in performance compared to bilinear fusion~\cite{gao2016compact}. Random Maclaurin projection and Tensor Sketch projection~\cite{gao2016compact} are the most prominent algorithms proposed for compact bilinear pooling. Here, we deploy the tensor sketch projection. This algorithm uses the count sketch projection introduced in~\cite{pham2013fast} to estimate the outer product of two vectors without computing the outer product explicitly. The count sketch of the outer product of two vectors can be expressed as the convolution of count sketches of the vectors~\cite{fukui2016multimodal}. However, this convolution can be computed as the inverse Fourier transform of the element-wise product of the count sketches in the frequency domain. Therefore, the bilinear outer product of multiple modalities can be computed through element-wise multiplication of Fourier domain count sketches. Let ${\bold  x_1}\in \mathbb{R}^{c_1}$ and ${ \bold x_2} \in \mathbb{R}^{c_2}$ be the modality-dedicated embedding layers:
{\small
\begin{equation}
{\bold y}={\text {FFT}}^{-1}({\text {FFT}}(\Psi({\bold x_1},h_1,s_1))\circ {\text { FFT}}(\Psi({\bold x_2},h_2,s_2))),
\end{equation}
}
where hash functions $h_1 \in \mathbb{N}^{c_1}$ and $h_2 \in \mathbb{N}^{c_2}$ are random, but fixed vectors uniformly drawn from $\{1,2,...,d\}$, $s_1\in \{-1,+1\}^{c_1}$, and $s_2\in \{-1,+1\}^{c_2}$. The count sketch function is defined as:
\begin{equation}
\Psi({\bold x_1},h_1,s_1))=\{(Qx_1)_1, (Qx_1)_2,...,(Qx_1)_d\},
\end{equation}
where $(Qx_1)_j=\sum_{n:h_1[n]=j}s_1[n]{\bold x_1}[n]$. This algorithm can be expanded to fuse multiple modalities as well.

In the proposed generalized compact bilinear fusion algorithm, single modalities and all possible $2$-,$3$-,$...,n$-compact bilinear products are concatenated to form vector $\bold y$. For instance, when $n=3$, three modality-dedicated embedding layer, three two-modality tensor sketch projection, and one three-modality tensor sketch projection are concatenated.\\  
{\bf End-to-end training of the architecture:} Generalized compact bilinear fusion algorithm  consists of random, but fixed functions $\{s_i\}$ and $\{h_i\}$, Fourier and inverse Fourier transforms. Since these transforms are differentiable, the error can be back-propagated through the fusion layer, the end-to-end training of the proposed generalized compact bilinear fusion algorithm is possible, and the multimodal architecture can be jointly optimized. For two-modality tensor sketch fusion algorithm, the error is back-propagated through the fusion layer using the equation below. Let $L$ represent the loss function at the fusion layer~\cite{gao2016compact}: 
\begin{equation}
\frac{\partial L}{\partial {\bold x_1}}=\sum_d \frac{\partial L}{\partial {\bold y}[d]} T_d^2({\bold x_2})\circ s_1,
\end{equation}
where ${T_d}^2({\bold x})\in \mathbb{R}^{c_1}$, ${T_d}^2({\bold x_2})[j]=\Psi({\bold x_2},h_2,s_2)[d-h_1[j]]$. 
Similarly, $\frac{\partial L}{\partial {\bold x_2}}$ can be calculated. 
\vspace{-3mm}
\section{JOINT OPTIMIZATION OF architecture}\label{sec:architecture}
\vspace{-3mm}
The multimodal CNN architecture consists of modality-dedicated CNN networks, a joint representation layer, and a softmax classification layer that are jointly trained and optimized. The modality-dedicated networks are trained to extract the modality specific features and the joint representation is trained to explore and enforce dependency between different modalities. The joint optimization of the networks, discards the unuseful features.\\
{\bf Modality-dedicated networks: }Each modality-dedicated CNN, consists of the first 16 layers of a conventional VGG19 network~\cite{simonyan2014very} and a fully-connected modality-dedicated embedding layer (FC6) of size $1024$. The fully-connected layers of the conventional VGG19 network are not practical for our application, since the joint optimization of the modality-dedicated networks and the joint representation layer is practically impossible due to the massive number of parameters that need to be trained and the limited number of training samples. 
The details for each modality-dedicated network can be found in Table~\ref{table:architecture2}.\\
{\bf Joint representation layer:} The output of the modality-dedicated networks are fused using one of the discussed fusion algorithm, 
then fed to a fully connected layer of size $1024$ and finally, fed to the softmax classification layer.
\begin{table}[t]
\small
\begin{center}
\addtolength{\tabcolsep}{-5pt}
\begin{tabular}{l@{\hskip .05in}c@{\hskip .05in}c@{\hskip .05in}c}
\toprule 
network & CNN-Face & CNN-Iris&CNN-Fingerprint\\
\hline
input&$224\times 224\times 3$&$512\times 64\times 3$&$224\times 224\times 3$\\
\toprule 
layer&kernel&kernel&kernel\\
\hline
conv1 (1-2)& $3\times 3 \times 64$ & $3\times 3 \times 64$& $3\times 3 \times 64$\\
\rowcolor{black!10} maxpool1 & 2 $\times 2$ &$ 2 \times 2$&$ 2 \times 2$\\
conv2 (1-2) & $3\times 3 \times 128$ & $3\times 3 \times 128$& $3\times 3 \times 128$\\
\rowcolor{black!10} maxpool2 & $2 \times 2$ & $2 \times 2$& $2 \times 2$\\
conv3 (1-4) & $3\times 3 \times 256$ & $3\times 3 \times 256$& $3\times 3 \times 256$\\
\rowcolor{black!10} maxpool3 & $2 \times 2$ & $2 \times 2$& $2 \times 2$\\
conv4 (1-4) & $3\times 3 \times 512$ & $3\times 3 \times 512$& $3\times 3 \times 512$\\
\rowcolor{black!10} maxpool4 & $2 \times 2 $& $2 \times 2$& $2 \times 2$\\
conv5 (1-4) & $3\times 3 \times 512$ & $3\times 3 \times 512$& $3\times 3 \times 512$\\
\rowcolor{black!10}FC6 &  $7\times 7\times 1024$ & $2\times 16\times 1024$&$7\times7\times 1024$\\
\bottomrule
\end{tabular}
\end{center}
\caption[Table caption text]{The modality-dedicated CNN architectures. 
}   
\label{table:architecture2}
\end{table}
\vspace{-3mm}
\section{Experiments and discussions}\label{sec:Experiments}
\vspace{-3mm}
{\bf  CMU Multi-PIE database}: This database~\cite{gross2010multi} consists of face images under different illuminations, viewpoints, and expressions which are recorded in four sessions. 
Following the setup in~\cite{bahrampour2016multimodal}, we consider the multi-view face images for $129$ subjects that are present in all sessions. The available views are divided into three modalities of $\{-90^\circ$, $-75^\circ$, $-60^\circ$, $-45^\circ\}$, $\{0^\circ$, $\pm 15^\circ$, $\pm 30^\circ\}$ and  $\{45^\circ$, $60^\circ$, $75^\circ$, $90^\circ\}$. Images from session 1 at views $\{0^\circ$, $\pm 30^\circ$, $\pm 60^\circ$,$\pm 90^\circ\}$ are used as training samples. Test images are obtained from all available view angles from session 2.  
\\
{\bf {BioCop multimodal database:}} This database~\cite{BIIC} is one of the few databases that allows disjoint training and testing of multimodal fusion at feature level. The BioCop database is collected under four disjoint years; 2008, 2009, 2012, and 2013. 
To make the training-test splits mutually exclusive, the 294 subject that are common in years 2012 and 2013 are considered. The proposed algorithm is trained on 294 mutual subjects in year 2013 dataset, and is tested on the same subjects in year 2012 dataset. It is worth mentioning that although the databases are labeled as 2012 and 2013, the date of data acquisition for common subjects in the datasets can vary between one to three years, which has also the advantage of investigating the effect of age-progression. We also consider the left and right irises as a single class, which results in heterogeneous classes for the iris modality. 
\\
{\bf BIOMDATA multimodal database}: This database~\cite{crihalmeanu2007protocol} is a challenging database, since many of the samples are damaged with blur, occlusion, sensor noise and shadows~\cite{haghighat2016discriminant}. 
Following the setup in~\cite{haghighat2016discriminant}, six biometric modalities are considered: left and right irises, and thumb and index fingerprints from both hands. The experiments are conducted on 219 subjects that have samples in all six modalities. For each modality, four randomly chosen samples are used for the training and the remaining samples are used for the test set. For any modality in which the number of the samples is less than five, one sample is used for the test set and the remaining samples are used for the  training. 
A summary of the databases is presented in Table~\ref{table:datasetsBIOMDATA}.\\ 
{\bf Training and test phases:} For each databases, the number of samples per individual and per modality varies. Therefore, for the training phase, for each individual $250$ sets of modalities are randomly chosen from the training set. Similarly  $250$ sets are chosen from test set for the test phase. For Multi-Pie and BioCop databases, each triplet includes one sample from each modality. Similarly, for BIOMDATA database each set includes normalized left and right irises, and enhanced left index, right index, left thumb and right thumb fingerprint images. For Multi-Pie database the number of triplets in training and test phases is the same and equal to $32,250$. The number of triplets in BioCop database and sets of six images in BIOMDATA database for training and test phase are equal to $73,500$ and $54,750$, respectively. 
\begin{table}[t]
\small
\begin{center}
\begin{tabular}{c| l c c c c c}
\toprule 
\multicolumn{1}{c}{} & \multicolumn{1}{c}{} & \multicolumn{1}{c}{Train set} & \multicolumn{1}{c}{Test set} &KNN&SVM&CNN\\ 
\hline
\multirow{3}{*}{\rotatebox[origin=c]{90}{BioCop}}&Face&6833&6960&89.68&88.76&98.14\\
&Iris&36636&39725&70.52&79.26&99.05\\
&Fingerprint&1822&991&91.22&90.61&97.28\\
\hline
\multirow{6}{*}{\rotatebox[origin=c]{90}{BIOMDATA}}&Left iris &874&584&66.61&71.92&99.35\\
&Right iris&871&581&64.89&71.08&98.95\\
&Left thumb&875&644&61.23&63.96 &80.15\\
&Left index&872&632&82.91&84.70 &93.43\\
&Right thumb&871&647&62.11&63.52&82.63\\
&Right Index&870&624&82.05&84.46&93.12\\
\hline
\multirow{3}{*}{\rotatebox[origin=c]{90}{Multi-Pie}}&Left view &10320&30940&45.52&47.30&87.50\\
&Frontal view&15480&38700&40.87&41.15&90.29\\
&Right view&10320&30960&45.13&47.30&85.49\\
\bottomrule
\end{tabular}
\end{center}
\caption[Table caption text]{The number of samples in training and test sets and rank-one recognition rate for single modalities.}
\label{table:datasetsBIOMDATA}
\end{table}
\\
{\bf Data representation:} The face images are cropped, aligned to a template~\cite{zhu2012face,dlib09}, and resized to $224\times 224$ images. Iris images are segmented, normalized using OSIRIS~\cite{krichen2008osiris}, and transformed into $64\times 512$ strips. Each fingerprint image is enhanced using the method described in~\cite{chikkerur2004systematic}, The core point is detected from the enhanced image~\cite{jain2000filterbank}, and finally a $224 \times 224$ region centered by the core point is cropped.
\\
{\bf Implementation:} Initially, each modality-dedicated CNNs is trained independently, and each CNN is optimized on a single modality. For each modality, the conventional VGG19 network is pre-trained on Imagenet~\cite{deng2009imagenet}. Pre-training helps with additional training data when the number of domain specific training data is limited. For the CNN-Face networks, the network is fine-tuned on CASIA-Webface~\cite{yi2014learning} and the corresponding database (BioCop 2013 or CMU Multi-Pie databases). The preprocessing algorithm includes the channel-wise mean subtraction on RGB values, where the channel means are calculated on the whole training set. 
\begin{table}[t]
\small
\begin{center}
\addtolength{\tabcolsep}{-0pt}
\begin{tabular}{lcccc}
\toprule 
Modality &  $\{$1,2$\}$&$\{$1,3$\}$&$\{$2,3$\}$&$\{$1,2,3$\}$\\
\hline
SVM-Major  &53.18&54.47&57.61&62.95 \\
SVM-Sum   &51.15&53.84&55.43&69.30           \\
SMDL &71.65&74.14&70.27& 81.30\\
JSRC   &68.16&66.42&64.53&73.30 \\
\hline
CNN-Major &92.18&93.75&89.74&95.87 \\
CNN-Sum&91.58&93.28&89.13&94.51\\
Weighted feature fusion&94.12&94.96&91.53&96.59\\
Generalized compact bilinear&94.67&95.53&92.18&97.27\\
\bottomrule
\end{tabular}
\end{center}
\caption[Table caption text]{Accuracy evaluation for different fusion settings for Multi-PIE database. 1, 2 and 3 represent frontal, right, and left views, respectively.}       
\label{table:results_CMUPie}
\end{table}
CNN-Iris networks are fine-tuned on CASIA-Iris-Thousand \cite{CASIA-IRIS}, Notre Dame-IRIS 04-05~\cite{bowyer2016nd}, and finally the corresponding database (BioCop-Iris 2013 or BIOMDATA database). For the BioCop database, the CNN-Fingerprint network is fine-tuned on the BioCop 2013 right index fingerprint database. For the BIOMDATA database, the networks are fine-tuned on the corresponding fingerprint databases. 

A two-step optimization algorithm is utilized to train the joint optimization of networks, where initially the modality-dedicated networks' weights are frozen and the joint representation layer is optimized greedily upon the extracted features by modality-dedicated networks. Then, all modality-dedicated networks, fusion layer, and the classification layer are jointly optimized. 
\\
{\bf Comparison of methods:} To compare the results for the proposed algorithms, with the state-of-the-art algorithms, Gabor features in five scales and eight orientations are extracted from all modalities. For each face, iris, and fingerprint image, $31,360$, $36,630$, and $31,360$ features are extracted respectively. These features are
used for all the algorithms except CNN-Sum, CNN-Major, and two proposed algorithms. Table~\ref{table:datasetsBIOMDATA} presents the results for the rank-one recognition rate for the databases. The performance of the proposed fusion algorithms is compared with several state-of-the-art feature, score and decision level fusion algorithms.  SVM-Sum and CNN-Sum use the probability outputs for the test sample of each modality, added together to give the final score vector. SVM-Major and CNN-Major chose the maximum number of modalities taken to be from the correct class. The feature level fusion techniques include serial feature fusion~\cite{liu2001shape}, parallel feature fusion~\cite{yang2003feature}, CCA-based feature fusion~\cite{sun2005new}, JSRC~\cite{shekhar2014joint}, SMDL~\cite{bahrampour2016multimodal}, and DCA/MDCA~\cite{haghighat2016discriminant} methods. 
Tables~\ref{table:results_CMUPie} and~\ref{table:results} present the results for different fusion settings. For all the databases we have considered $d=4096$. For BIOMDATA database, due to the vast number of possible outer products, the generalized compact bilinear method only includes single modalities and three compact bilinear multiplications (two irises, two index fingers and two thumbs). 
The reported values are the average values for five randomly generated training and test sets for the training and test phases.
\begin{table}[t]
\small
\begin{subtable}[h]{0.5\textwidth}
\begin{center}
\addtolength{\tabcolsep}{-0pt}
\begin{tabular}{lccccc}
\toprule 
Modality & $\{$1,2$\}$&$\{$1,3$\}$&$\{$2,3$\}$&$\{$1,2,3$\}$\\
\hline
SVM-Major & 79.22 & 89.27&80.47 &90.32\\
Serial + PCA + KNN&71.12& 86.28&75.69&76.18\\
Serial + LDA + KNN&80.12&91.28&79.69&82.18 \\
Parallel + PCA + KNN&74.69&88.12&77.58&-\\
Parallel + LDA + KNN&82.53&93.21&82.56&- \\
CCA + PCA + KNN&87.21&95.27&86.44&95.33\\
CCA + LDA + KNN&89.12&95.41&86.11&95.58\\
DCA/MDCA + KNN&83.02&96.36&83.44&86.49\\
\hline
CNN-Sum &99.10 & 98.85& 98.92&99.14\\
CNN-Major &98.51&97.70&98.31&99.03\\
Weighted feature fusion& 99.18&99.03&99.12&99.25\\
Generalized compact bilinear&99.27&99.12&99.16&99.30\\
\bottomrule
\end{tabular}
\end{center}
\caption[Table caption text]{BioCop database: 1, 2, and 3 represent face, iris, and fingerprint, respectively.}   
\label{table:results_Biocop}
\end{subtable}
\begin{subtable}[h]{0.5\textwidth}
\begin{center}
\addtolength{\tabcolsep}{-0pt}
\begin{tabular}{lccc}
\toprule 
Modality & 2 irises & 4 fingerprints& 6 modalities\\
\hline
SVM-Major &78.12  &88.34 &93.31 \\
SVM-Sum   &81.23      &94.13     &96.85     \\
Serial + PCA+ KNN &72.31&90.71&89.11 \\
Serial + LDA+ KNN&79.82&92.62&92.81 \\
Parallel + PCA+ KNN&76.45&-&-\\
Parallel + LDA+ KNN&83.17&-&-\\
CCA + PCA + KNN&88.47&94.72&94.81\\
CCA + LDA + KNN&90.96&94.13&95.12 \\
JSRC&78.20&97.60&98.60\\
SMDL&83.77&97.56&99.10\\
DCA/MDCA + KNN&83.77&98.1&99.60\\
\hline
CNN-Sum&99.54&99.46&99.82\\
CNN-Major&99.31&99.42&99.48\\
Weighted feature fusion&99.73&99.65&99.86\\
Generalized compact bilinear&99.79&99.70&99.90\\
\bottomrule
\end{tabular}
\end{center}
\caption[Table caption text]{BIOMDATA database.}       
\label{table:results_BIOMDATA}
\end{subtable}
\caption[Table caption text]{Accuracy evaluation for different fusion settings.}       
\label{table:results}
\end{table}
\vspace{-7mm}
\section{Conclusion}
\vspace{-3mm}
In this paper, we proposed a joint CNN architecture with feature level fusion for multimodal recognition using multiple modalities. We proposed to apply fusion at fully-connected layers instead of convolutional layers to handle the possible spatial mismatch problem. This fusion algorithm results in no loss in performance, while the number of parameters is reduced significantly. We demonstrated that the multimodal fusion at the feature level and joint optimization of multi-stream CNNs significantly improve unimodal representation accuracy by incorporating the captured multiplicative interactions of the low-dimensional modality-dedicated feature representations, by means of generalized compact bilinear pooling.
\begin{center}
ACKNOWLEDGEMENT
\end{center}
\vspace{-3mm}
This work is based upon a work supported by the Center for Identification Technology Research and the National Science Foundation under Grant $\#1650474$.

{\footnotesize 	
\bibliographystyle{IEEEtran}
\bibliography{bib}
}


\end{document}